\documentclass{article}

\usepackage[nonatbib,final]{nips_OptOpt}

\usepackage{verbatim}

\usepackage{ifpdf}
\usepackage{cite}
\usepackage[pdftex]{graphicx}
\usepackage{amsmath,amssymb,amsthm,amsfonts}
\usepackage{algorithm}
\usepackage{algorithmic}
\usepackage{array}
\usepackage{url}
\usepackage{color}

\newtheorem{theorem}{Theorem}

\newtheorem{assumption}[theorem]{Assumption}

\newtheorem{corollary}[theorem]{Corollary}

\DeclareMathOperator*{\argmin}{arg\,min}

\newcommand{\algoref}[1]{Algorithm~\ref{#1}}

\newcommand{\thref}[1]{Theorem~\ref{#1}}

\begin{document}

\title{Tuning Over-Relaxed ADMM}

\author{
Guilherme Fran\c{c}a \\
Johns Hopkins University\\
\texttt{guifranca@gmail.com} \\
\And
Jos\'e Bento \\
Boston College\\
\texttt{jose.bento@bc.edu}
}

\maketitle

\begin{abstract}
The framework of Integral Quadratic Constraints (IQC)
reduces the computation
of \mbox{upper}
bounds on the convergence rate of several optimization algorithms
to a semi-definite program (SDP). In the case of over-relaxed
Alternating Direction Method of Multipliers (ADMM),
an explicit and  closed form solution to this SDP was
derived in our recent work \cite{FrancaBento}.
The purpose of this paper is twofold.
First, we summarize these results.
Second, we explore one of its consequences which allows us to obtain
general and simple formulas for optimal parameter selection.
These results are valid for arbitrary strongly convex objective functions.
\end{abstract}

\section{Introduction} \label{sec:introduction}

Consider the optimization problem
\begin{equation} \label{minimize}
\min_{x\in \mathbb{R}^p, z\in \mathbb{R}^q} \{ f(x) + g(z) \}
\quad  \text{subject to} \quad Ax + Bz = c ,
\end{equation}
where $A\in \mathbb{R}^{r \times p}$,
$B\in \mathbb{R}^{r \times q}$,
and $c\in \mathbb{R}^{r}$.
We consider
ADMM applied to problem \eqref{minimize} under the
assumption that $f(x)$ is strongly convex and $g(z)$ is convex.
ADMM is parametrized by $\alpha > 0$ and $\rho > 0$, and
takes the form of Algorithm~\ref{ADMM}.
Strictly speaking, this defines a family of
algorithms, one per parameter choice.

In this paper,
we tune ADMM by providing explicit and simple formulas for
the parameters $\alpha$ and $\rho$, yielding
the best possible asymptotic convergence rate among all first order methods.
This is an immediate consequence of the results proposed in \cite{FrancaBento}.

%

A classical choice of parameters is $\alpha = 1$ and $\rho = 1$, which can
be substantially suboptimal.
Several works have computed bounds on ADMM's convergence rate
for specific and restricted ranges of $\alpha$ and $\rho$.
However, the IQC formalism introduced in \cite{Lessard},
allowed \cite{Jordan} to reduce the analysis of this entire family
of solvers to finding solutions to an SDP.
This SDP has multiple solutions, each one giving a different
bound on the convergence rate of ADMM, some better than others.
This SDP was analyzed numerically,
and a single explicit feasible
solution was given when $\kappa$ is sufficiently large ($\kappa$  is
related to the ratio of the smallest to the largest ``curvature'' of $f$).
It was also shown via a lower bound, that for large $\kappa$,
it is not possible to extract from this SDP a rate much better than this.

A \emph{closed form solution} to this SDP was recently
obtained in \cite{FrancaBento},
which express the convergence rate \emph{explicitly} in
terms of the parameters of ADMM and condition numbers of problem
\eqref{minimize}.
Moreover, it was shown that the closed form solution is the best possible one can
extract from the SDP.
Here we revisit these results, and from this explicit solution we
provide formulas for the
optimal parameters $\alpha$ and $\rho$ of ADMM in terms of
condition numbers and curvature of $f$.

\begin{algorithm}[t]
\caption{
\label{ADMM}
Family of over-relaxed ADMM schemes
(parameters $\alpha$, $\rho$)
}
\begin{algorithmic}[1]
  \STATE {\bfseries Input:} $f$, $g$, $A$, $B$, $c$;
  \STATE Initialize $z_0, u_0$
  \REPEAT
  \STATE $x_{t+1} = \argmin_x f(x) + \tfrac{\rho}{2} \|Ax + Bz_t - c + u_t\|^2$
  \STATE $z_{t+1} = \argmin_z g(z) + \tfrac{\rho}{2} \|\alpha Ax_{t+1} - (1-\alpha)Bz_t  + Bz - \alpha c + u_t\|^2$
  \STATE $u_{t+1} = u_t + \alpha Ax_{t+1} - (1-\alpha)Bz_t + Bz_{t+1} - \alpha c$
  \UNTIL{stopping criteria}
\end{algorithmic}
\end{algorithm}

\section{Main Results}

\begin{assumption}\label{assumption}
Throughout the paper, we assume that $f$ and $g$ in \eqref{minimize}
are convex, closed and
proper, $A$ is invertible, and $B$ has full column rank.
Given a function $h: \mathbb{R}^p \mapsto \mathbb{R}$
we say that $h \in S_p(m,L)$ if and only if $0 < m \le L < \infty$ and
$m\|x-y\|^2\le\left(\nabla h(x)-\nabla h(y)\right)^T(x-y)\le L\| x-y\|^2$.
In other words,
$S_p(m,L)$ is the set of strongly convex functions with Lipschitz
continuous gradients. We assume that
$f \in S_p(m,L)$ and $g \in S_q(0,\infty)$.
\end{assumption}

We start by recalling the main result of \cite{Jordan}.
It was shown that the iterative scheme of
\algoref{ADMM} can be written as a dynamical system with a feedback
signal related to the gradient of $f$ and sub-gradient of $g$.
The stability of this dynamical system is then related to the
convergence rate of \algoref{ADMM}, which in turn involves
numerically solving a $4\times4$ SDP,
as stated below in Theorem~\ref{jordan_theo}.
We state this in a simplified form, and refer the reader
to \cite{Jordan} for more details.
Let us first introduce the constants
\begin{equation}\label{parameters}
\rho_0 = \rho \,(\hat{m}\hat{L})^{-1/2}, \qquad
\kappa = \kappa_f \kappa_A^2
\end{equation}
where $\hat{m} = {m}/{\sigma_1^2(A)}$,
$\hat{L} = {L}/{\sigma_p^2(A)}$,
and $\kappa_f = L/m$. Here $\sigma_1(A)$ and $\sigma_p(A)$ denote
the largest and smallest singular
value of matrix $A$, respectively.
In addition, $\kappa_A = \sigma_1(A)/\sigma_p(A)$ is the condition
number of $A$.

\begin{theorem}[See \cite{Jordan}]\label{jordan_theo}
Let the sequences
$\left\{ x_t \right\}$,
$\left\{ z_t \right\}$, and
$\left\{ u_t \right\}$ evolve according to \algoref{ADMM}.
Let $\varphi_t = \left[ z_t, u_t\right]^{T}$
and $\varphi_{*}$ be a fixed point.
Let $0 < \tau < 1$ be such that
\begin{equation}\label{semidefinite}
E - \tau^2 F + G\preceq 0,
\end{equation}
where $E$, $F$ and $G$ are $4 \times 4$
matrices depending on $\alpha$, $\rho_0$,
and $\kappa$. Moreover, $G$ is symmetric,
$E$ is positive semi-definite,
and $F = \left( \begin{smallmatrix} P & 0 \\ 0 & 0\end{smallmatrix}\right)$
where $P$ is a $2 \times 2$ positive-definite matrix
(the exact definitions are not important for this paper).
For all $t\ge 0$ we thus have
\begin{equation}\label{bound_admm}
\| \varphi_t - \varphi_* \| \le \kappa_B \sqrt{\kappa_P} \, \tau^t
\| \varphi_0 - \varphi_* \|
.
\end{equation}
\end{theorem}

As already pointed out in \cite{Jordan}, the weakness of
\thref{jordan_theo} is
that $\tau$ is not explicitly given as a function of
$\kappa$, $\rho_0$, and $\alpha$. The factor
$\kappa_P$ in \eqref{bound_admm} is also not explicitly given. Therefore,
for given values of $\kappa$, $\rho_0$, and $\alpha$, one must perform
a numerical search to find the minimal $\tau$ such
that \eqref{semidefinite} is feasible, and thus obtain the best possible bound
on the convergence rate of ADMM. Notice, however,
that it is unclear a priori whether other methods could improve
on this bound. This numerical approach was carried out in \cite{Jordan}
using a binary search on $\tau$, justified by the fact that the positive
semi-definite property of $F$ implies that the eigenvalues
of $E - \tau^2 G + G$ decrease monotonically with $\tau$. Notice that
one might have to scan the parameter space $(\alpha, \rho_0)$ multiple times
for a problem with a specific value of $\kappa$.
Even from a practical point of view, this procedure may
introduce delays. For instance, if \eqref{semidefinite}
is used in an adaptive scheme where after every few iterations we
estimate a local value of $\kappa$ and then re-optimize $\alpha$ and $\rho$.

Therefore, it is not only theoretically desirable to have an explicit
expression for the smallest
$\tau$ that \eqref{semidefinite} can provide, but it may also be useful
in  practical applications. Such result was
proposed in \cite{FrancaBento}, and reproduced below in
Theorem~\ref{explicit_solution}. Let us first introduce the function
\begin{equation}
\chi(x) = \max(x, x^{-1}) \ge 1 \quad \mbox{for $x \in \mathbb{R} > 0$}.
\end{equation}

\begin{theorem}[See \cite{FrancaBento}] \label{explicit_solution}
Let $0 < \alpha < 2$, $\kappa \geq 1$, and $\rho_0>0$.
The convergence rate of Algorithm~\ref{ADMM} satisfies
\begin{equation} \label{eq:full_upper_bound_admm}
\| \varphi_t - \varphi_* \| \le \kappa_B \sqrt{\chi(\eta)} \, \tau_A^{\, t}
\| \varphi_0 - \varphi_* \|
\end{equation}
where
\begin{equation}\label{tau_sol}
\tau_A = 1 - \dfrac{\alpha}{1+\chi(\rho_0)\sqrt{\kappa}} \quad \mbox{and}
\quad
\eta = \dfrac{\alpha}{2-\alpha} \cdot
\dfrac{\chi(\rho_0)\sqrt{\kappa}-1}{\chi(\rho_0)\sqrt{\kappa}+1}.
\end{equation}
Moreover, $\tau_A$ in \eqref{tau_sol} is the smallest
possible $\tau$ which solves \eqref{semidefinite}.
\end{theorem}

No other proof strategy can give a better general
upper bound than \eqref{tau_sol}
since $\tau_A$ is actually \emph{attainable}. For instance, choosing
$f(x) = \tfrac{1}{2} x^T Q x$, where
$Q = \text{diag}(m,L)$, $g(z) = 0$, $A = I$, $B = - I$, and $c = 0$,
the convergence rate of ADMM is given \emph{exactly} by the formula
of $\tau_A$ in \eqref{tau_sol} for
any values of $\kappa > 1$, $0 < \alpha < 2$ and $\rho_0>0$.

As a consequence of Theorem~\ref{explicit_solution}, we now present
an explicit formula for optimal parameter selection of
Algorithm~\ref{ADMM}.
\begin{corollary}[Optimal Parameter Selection]\label{optimal_parameters}
The best asymptotic convergence rate of over-relaxed ADMM,
for $\alpha \in (0,2)$ and $\rho \in (0,\infty)$, is given by
\begin{equation} \label{inf_tau}
\inf_{ \alpha, \rho  } \tau_{A}
= 1 - \frac{2}{1 +  \sqrt{\kappa}} ,
\end{equation}
where, we recall, $\kappa =
\big(\tfrac{\sigma_1(A)}{\sigma_p(A)}\big)^2 \tfrac{L}{m}$,
and is achieved by letting
$\alpha \to 2^{-}$ and
$\rho = \tfrac{\sqrt{mL}}{\sigma_1(A) \sigma_p(A)}$.
Furthermore, for a fixed iteration number $t$,
the upper bound \eqref{eq:full_upper_bound_admm} is minimized by
\begin{equation}\label{best_alpha}
\alpha = \begin{cases}
1 + \dfrac{1}{\chi(\rho_0)\sqrt{\kappa}} & \quad \mbox{if
$t \leq \chi(\rho_0) \sqrt{\kappa}$,} \\
1+ \dfrac{1+ \sqrt{1+4t^2 - 4t\chi(\rho_0)\sqrt{\kappa} }}{2t} & \quad
\mbox{if $t > \chi(\rho_0) \sqrt{\kappa}$.}
\end{cases}
\end{equation}
\end{corollary}
Note that the optimal $\alpha$ and $\rho$ in
Corollary~\ref{optimal_parameters} are expressed only in terms of
condition numbers of problem \eqref{minimize}, i.e.
singular values of $A$ and bounds on the curvature of $f$.
The matrix $B$ affects convergence but not the optimal choice of parameters.
The function $g$ does not affect the bound on the convergence rate.
The above  tuning rule optimizes
a \emph{general} bound that holds simultaneously for all problems with the same
$\kappa$. There is no tighter bound than this for the same general setting.
However, this does not mean that for a specific problem
a better bound with different tuning parameters cannot be found.

\paragraph{Related Work.}
Two of the most explicit bounds that resemble \eqref{tau_sol}
are found in \cite{boyd_almost_bound,pontus} and \cite{wei_almost_bound}.
In \cite{boyd_almost_bound,pontus} the Douglas-Rachford splitting method
is analyzed, which is different but related to the scheme considered
in this paper. For a problem similar to \eqref{minimize}, it gives
a rate bound of $1 - {\alpha}/{(1 + \sqrt{\kappa_f})}$,
where $\alpha$ is a step size and $\kappa_f = L/m$.
ADMM to problem \eqref{minimize} with
$\alpha =1$ and
$\rho_0 = 1$ is considered in \cite{wei_almost_bound},
and give an
approximate rate bound of
$1 - {1}/{\sqrt{\kappa}} + O({\kappa^{-1}})$, where
$\kappa = \kappa_f \kappa^2_A$.
There are other works on ADMM with exact bound calculations.
For example, \cite{iutzeler2016explicit} focus on distributed ADMM
but only for the non-relaxed version.
Explicit convergence rate and optimal parameters
are given in \cite{Ghadimi}, however, only for the particular case of
quadratic objectives.
Similar expressions for the convergence rate were proposed in
in \cite{shi2014linear}.
These expressions are upper bounds which are optimized, but it is not
possible to prove
they are the best possible.
In addition, this work do not focus on over-relaxed ADMM.
Finally, \cite{boley2013local} and \cite{davis2014faster}
also study over-relaxed ADMM. They provide upper bounds, but again,
cannot prove these are the best possible.
In \cite{davis2014faster} one can find a table that gives a good
summary of known bounds under different assumptions,
none of which overlaps with our work.

\section{Numerical results}

\begin{figure}
\centering
\includegraphics[width=.3\linewidth]{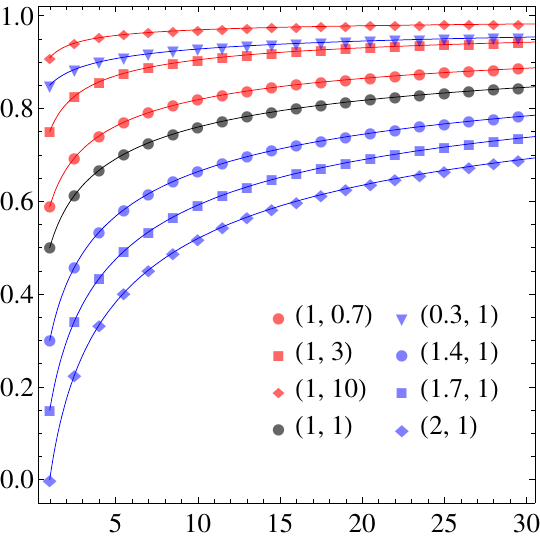}
\put(-60,120){\small (a)}
\put(-60,-7){\small $\kappa$}
\put(-130,70){\small $\tau$}
\put(-60,60){\footnotesize $(\alpha,\rho_0)$}
\hspace{0.5cm}
\includegraphics[width=.3\linewidth]{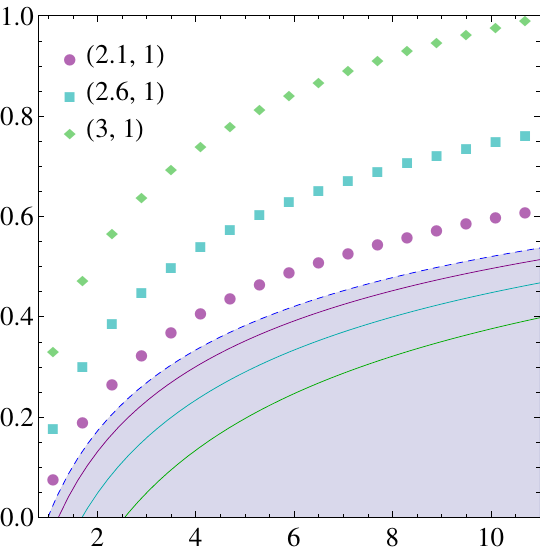}
\put(-60,120){\small (b)}
\put(-55,-7){\small $\kappa$}
\put(-130,70){\small $\tau$}
\put(-80,105){\footnotesize $(\alpha,\rho_0)$}
\hspace{0.5cm}
\includegraphics[width=.315\linewidth]{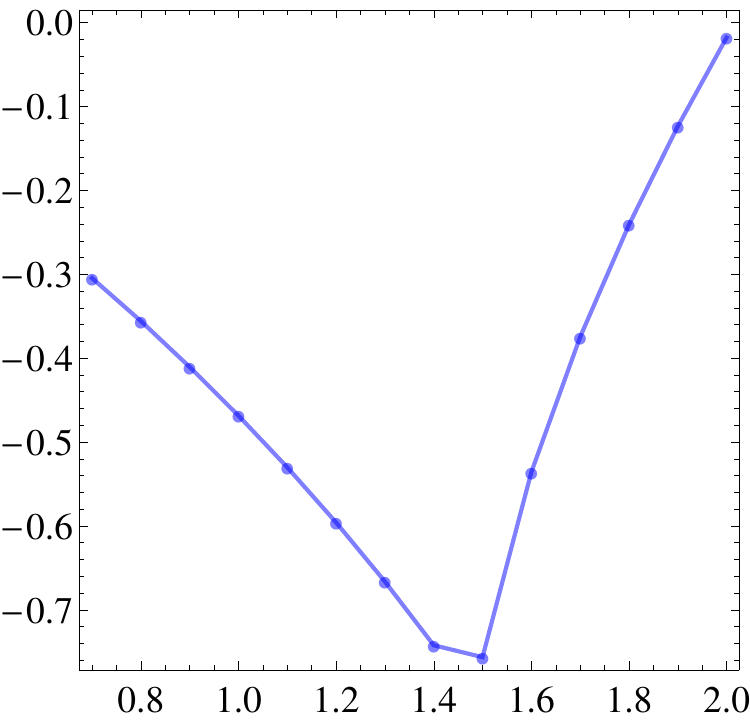}
\put(-60,120){\small (c)}
\put(-55,-7){\small $\alpha$}
\put(-100,70){\small $\log(\tau)$}
\caption{ \label{tau_numerical} \label{alpha_min} \label{alpha_bigger}
(a) Numeric (dots) and exact (lines) plot of $\tau$
versus $\kappa$ for different values of
$\alpha$ and $\rho_0$.
The best choice of parameters is $\rho_0=1$ and $\alpha=2$.
(b) Numeric (dots) and exact (lines) plot of $\tau$ versus $\kappa$ with $\alpha > 2$ and $\rho_0=1$.
The dashed blue line corresponds to $\alpha=2$ in
$\tau$. The shaded region contains curves $\tau$
for values of $\alpha$ not allowed
in \thref{explicit_solution}.
Numerical solutions with $\alpha > 2$ are restricted
$1 < \kappa \lesssim 11$. Notice that $\alpha > 2$ does not produce
better convergence rates than $1\le\alpha<2$ through \eqref{tau_sol}.
(c) Numeric $\log \tau$ for different values of $\alpha$ when solving a classification problem.
}
\end{figure}
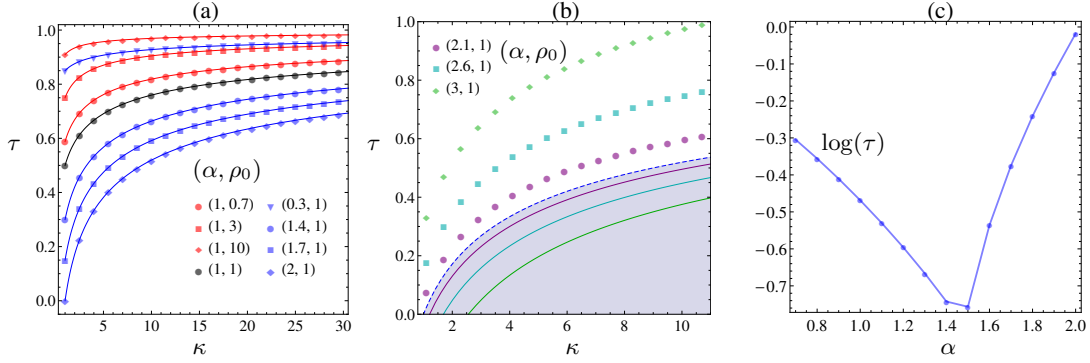

Let us compare numerical solutions to the SDP in \thref{jordan_theo}
with the exact $\tau_A$ in equation \eqref{tau_sol}.
We use a binary search to find the best $\tau$ that solves
\eqref{semidefinite}.
Figure~\ref{tau_numerical}~(a) shows the rate bound $\tau$ against $\kappa$
for several choices of parameters $(\alpha,\,\rho_0)$.
The dots correspond to the numerical solutions
and the solid lines correspond to the exact formula $\tau_A$ in
\eqref{tau_sol}.

Theorem \ref{explicit_solution} is valid only
for $0<\alpha <  2$ ($\tau_A$ can assume negative values
for $\alpha > 2$). However,
\thref{jordan_theo} does not impose any
restriction on $\alpha$, and holds even for $\alpha > 2$ \cite{Jordan}.
To explore the range $\alpha > 2$,
we numerically solve \eqref{semidefinite} as
shown in Figure~\ref{alpha_bigger}~(b).
The dots correspond to the numerical solutions.
The dashed blue line corresponds to \eqref{tau_sol} with
$\alpha=2$, and it is the boundary of the shaded region in which
\eqref{tau_sol} can have negative values and is no longer
valid. Although \thref{explicit_solution} does not hold for $\alpha>2$,
we deliberately
included the solid lines representing \eqref{tau_sol} inside
this region.  Obviously, these curves do not match the numerical results.

The first important remark is that, for a given $\alpha > 2$, we were
unable to numerically find solutions for arbitrary $\kappa \geq 1$.
For instance, for $\alpha=2.6$ we can only stay roughly on
the interval $1<\kappa \lesssim11$.
The same behavior occurs for any \mbox{$\alpha > 2$}, and the range of
$\kappa$ becomes narrower as $\alpha$ increases.
From the picture one can notice that $\tau = 1$ is actually attained with
\emph{finite} $\kappa$, while for \eqref{tau_sol} this never happens;
it rather approaches $\tau\to 1^{-}$ as $\kappa \to \infty$.
Therefore, although it is feasible to solve \eqref{semidefinite}
with $\alpha > 2$, the solutions will be constrained to a small range
of $\kappa$. The next question would be if \thref{jordan_theo} for
$\alpha > 2$ could possibly
give a better rate bound than
Theorem \ref{explicit_solution} with $1 \le \alpha < 2$. We
can see from the picture that this is probably not the case.

Now let us consider problem
$\min_{\theta \in \mathbb{R}^d} \{ f(\theta) + g(\theta) \}$
for the following
regularized logistic regression to learn a sparse classifier from $N$ pairs $(x_i,y_i)$ where $x \in \mathbb{R}^d$ are features and $y \in \{-1,+1\}$ labels:
%
\begin{equation}
f(\theta) =  \dfrac{1}{N} \sum_{i=1}^N
\log\left(1 + e^{-y_i \theta^T x_i}\right), \qquad
g(\theta) = \mathbb{I}_\infty( \| \theta \|_1 > \lambda),
\end{equation}
where $\mathbb{I}_\infty(\bullet) = 0$ if $(\bullet)$ is
false and $\infty$ otherwise. In \eqref{minimize} we have $A = I$ and $B = -I$
so $\kappa_A = \kappa_B = 1$. We generate $N/2$ points with $y_i = +1$ and $N/2$ points with $y_i = -1$
from a gaussian distribution $\mathcal{N}(0,\sigma I)$
in even $d$ dimensions. Then,
for the $+1$ points, we shift half of the features by $+1/2$,
and for the $-1$ points we shift the same half of the features by $-1/2$.
Thus we have two classes of points whose centers
are separated by $1$ along $d/2$ dimensions and the best hyper-plane
separating these two classes has sparse coefficients. We use
Algorithm~\ref{ADMM} to solve this problem with several values of
$\alpha$ and $\rho$, and plot $\log\left( \| \theta_t - \theta^\star \|\right)$ against
iteration number $t$, where $\theta^\star$ is the optimum.
If $N \geq d$, the function $f$ restricted to
$\|\theta\|_1 \leq \lambda$ is strongly convex with high
probability.
If $\lambda$ is small then $\|\theta\|$ is small and so
$\nabla^2 f(\theta) = \tfrac{1}{N} \sum^N_{i
= 1} \tfrac{x_i x^T_i}{\cosh(x^T_i \theta /2)}
\approx \tfrac{1}{N} \sum^N_{i = 1} x_i x^T_i$.
From this, even without knowing $\theta$,
we can estimate $\kappa_f$ which,
for our choice of $N$, $d$, $\sigma$ and $\lambda$,
is very large. Hence, our estimate for the
best $\alpha$ using \eqref{best_alpha} is $1$.
We see from Figure~\ref{alpha_min}~(c) that this
is not the best choice, which is $\alpha \approx 1.5$.
Recall that our tuning rule is the best that holds uniformly across the
family of strongly convex functions, but for specific problems
it might be suboptimal.

\vspace{-0.25cm}
\section{Conclusion}
\vspace{-0.25cm}

We summarized the main results of \cite{FrancaBento}, which
is the content of Theorem~\ref{explicit_solution}.
This introduces a new and explicit upper bound on the convergence
rate of the family of over-relaxed ADMM, for arbitrary but
strongly convex objective functions.
This improves on
previous work \cite{Jordan,wei_almost_bound}.
In particular, the only explicit bound in \cite{Jordan} is a special
case of \eqref{tau_sol} when $\kappa$ is large.
Moreover, \eqref{tau_sol} is the best one can extract from the
IQC framework of \cite{Lessard}.

From this general bound we provide a tuning scheme for ADMM; Corollary \ref{optimal_parameters}.
In \cite{Nesterov} we find that
$1 - 2 / (1 + \sqrt{\kappa})$, where $\kappa = m/L$,
bounds the convergence rate of any first order method on
$S_p(m,L)$. Thus the ADMM tuned as in \eqref{inf_tau} is close to optimal
as a scheme for the entire family of strongly convex functions. However, as
shown
in the numerical experiments,
for specific problems our tuning might be suboptimal.






\begin{thebibliography}{10}

\bibitem{FrancaBento} G. Fran\c ca, J. Bento,
``An Explicit Rate Bound for the Over-Relaxed ADMM'',
\textit{ISIT} (2016),
arXiv:1512.02063v2 [stat.ML]

\bibitem{Lessard} L. Lessard, B. Recht, A. Packard,
``Analysis and design of optimization algorithms via integral
quadratic constraints'' (2014), arXiv:1408.3595 [math.OC]

\bibitem{Jordan} R. Nishihara, L. Lessard, B. Recht, A. Packard,
M. I. Jordan, ``A General Analysis of the Convergence of ADMM'',
\textit{Int. Conf. on Machine Learning} 32 (2015),
arXiv:1502.02009 [math.OC]


\bibitem{Ghadimi} E. Ghadimi, A. Teixeira, I. Shames,
``Optimal Parameter Selection for the Alternating Direction
Method of Multipliers (ADMM): Quadratic Problems'',
\textit{IEEE Trans. on Automatic Control} 60 4 (2015)

\bibitem{Nesterov} Y. Nesterov, ``Introductory Lectures on Convex
Optimization: A Basic Course'', Kluwer Academic Publishers, Boston, MA,
2004

\bibitem{boyd_almost_bound} P. Giselsson, S. Boyd.
``Diagonal scaling in Douglas-Rachford splitting and ADMM'',
\textit{Decision and Control (CDC)},
(2014) IEEE 53rd Annual Conference

\bibitem{pontus} P. Giselsson, S. Boyd.
``Linear Convergence and Metric Selection for Douglas-Rachford Splitting
and ADMM'', \textit{IEEE Transactions on Automatic Control}, (2016): 62

\bibitem{wei_almost_bound} D. Wei, W. Yin.
``On the global and linear convergence of the
generalized alternating direction method of
multipliers'', \textit{Journal of Scientific Computing} (2012): 1-28

\bibitem{iutzeler2016explicit} F. Iutzeler, P. Bianchi, P. Ciblat and W. Hachem.
``Explicit convergence rate of a distributed alternating direction
method of multipliers'',
\textit{IEEE Transactions on Automatic Control} (2016): 61

\bibitem{shi2014linear} W. Shi,  Q. Ling, K. Yuan, G. Wu and W. Yin.
``On the linear convergence of the ADMM in decentralized
consensus optimization'',
\textit{IEEE Transactions on Signal Processing} (2014): 62

\bibitem{boley2013local} D. Boley.
``Local linear convergence of the alternating
direction method of multipliers on
quadratic or linear programs'',
\textit{SIAM Journal on Optimization} (2013): 23

\bibitem{davis2014faster} D. Davis and W. Yin.
``Faster convergence rates of relaxed Peaceman-Rachford and
ADMM under regularity assumptions'',
arXiv:1407.5210 [math.OC] (2014)


\end{thebibliography}
%

\end{document}